\pdfoutput=1
\documentclass[11pt]{article}

\usepackage{emnlp2021}

\usepackage{times}
\usepackage{latexsym}

\usepackage[T1]{fontenc}
\usepackage[utf8]{inputenc}

\usepackage{microtype}
\usepackage{svg}
\usepackage{amsmath,amssymb}
\usepackage{here}
\usepackage{url}
\usepackage{comment}
\usepackage{booktabs}
\usepackage{multirow}
\usepackage{tabularx}
\usepackage{soul}
\usepackage{pifont}
\usepackage{tcolorbox}
\usepackage{CJK}
\usepackage{colortbl}

\newcommand{\squaddu}[1]{SQuAD$^{\rm Du}_{\rm #1}$}
\newcommand{\cmark}{\ding{51}}%
\newcommand{\xmark}{\ding{55}}%

\title{Can Question Generation Debias Question Answering Models?\\
A Case Study on Question--Context Lexical Overlap}

\author{Kazutoshi Shinoda$^{1,2}$ ~~ Saku Sugawara$^2$ ~~ Akiko Aizawa$^{1,2}$\\
    $^1$The University of Tokyo\\
    $^2$National Institute of Informatics\\
    {\tt shinoda@is.s.u-tokyo.ac.jp} \\
    {\tt \{saku,aizawa\}@nii.ac.jp} \\
}

\begin{document}
\maketitle
\begin{abstract}
Question answering (QA) models for reading comprehension have been demonstrated to exploit unintended dataset biases such as question--context lexical overlap.
This hinders QA models from generalizing to under-represented samples such as questions with low lexical overlap.
Question generation (QG), a method for augmenting QA datasets, can be a solution for such performance degradation if QG can properly debias QA datasets.
However, we discover that recent neural QG models are biased towards generating questions with high lexical overlap, which can amplify the dataset bias.
Moreover, our analysis reveals that data augmentation with these QG models frequently impairs the performance on questions with low lexical overlap, while improving that on questions with high lexical overlap.
To address this problem, we use a synonym replacement-based approach to augment questions with low lexical overlap.
We demonstrate that the proposed data augmentation approach is simple yet effective to mitigate the degradation problem with only 70k synthetic examples.\footnote{Our data is publicly available at \url{https://github.com/KazutoshiShinoda/Synonym-Replacement}.}
\end{abstract}

\begin{table*}[htbp]
\setlength{\tabcolsep}{2pt}
\centering\small
\setlength{\tabcolsep}{0.1cm}
\begin{tabular}{p{3.8cm}p{1.4cm}cp{1.9cm}p{3.8cm}c}
\toprule
\multicolumn{6}{c}{$\boldsymbol{C}$}\\
\midrule
\multicolumn{6}{p{15cm}}{Besides earning a reputation as a respected entertainment device, the iPod has also been accepted as a business device. Government departments, major institutions and international organisations have turned to the iPod line as a delivery mechanism for business communication and training, such as the Royal and Western Infirmaries in Glasgow, Scotland, where iPods are used to train new staff.}\\
\midrule
\multicolumn{1}{c}{$\boldsymbol{Q}$} & \multicolumn{1}{c}{$\boldsymbol{A}$} &
\multicolumn{1}{c}{$\boldsymbol{\frac{|Q \cap C|}{|Q|}}$} & 
\multicolumn{1}{c}{$\boldsymbol{C, Q \rightarrow A'}$} & \multicolumn{1}{c}{$\boldsymbol{C, A \rightarrow Q'}$}
& \multicolumn{1}{c}{$\boldsymbol{\frac{|Q' \cap C|}{|Q'|}}$}\\
\midrule
\ul{Where} is \ul{Royal} \ul{and} \ul{Western} \ul{Infirmaries} located?
&
Glasgow,
Scotland
&
5/8 = \colorbox{green!60}{0.62} &
Glasgow,
Scotland (\cmark)
&
\ul{Where} is \ul{the} \ul{Royal} \ul{and} \ul{Western} \ul{Infirmaries} located? (\cmark)
&
6/9 = \colorbox{green!60}{0.67}
\vspace{0.1cm}
\\ \midrule
Aside from recreational use\ul{,} \ul{in} what other arena \ul{have} \ul{iPods} found use?
&
 \centering business & 
4/14 = \colorbox{red!50}{0.29} &
\centering entertainment (\xmark) &
\ul{The} \ul{iPod} \ul{has} \ul{been} \ul{accepted} \ul{as} what kind of \ul{device}? (\xmark)
&
7/11 = \colorbox{green!60}{0.64}
\\
\bottomrule
\end{tabular}
\caption{Examples of ground-truth question--answer pairs and predictions of question answering (BERT-base \cite{bert}) and generation (SemanticQG \cite{zhang-bansal-2019-addressing}) models.
$\boldsymbol{C}$: context,
$\boldsymbol{Q}$: question,
$\boldsymbol{A}$: answer,
$\boldsymbol{\frac{|Q \cap C|}{|Q|}}$: question--context lexical overlap,
$\boldsymbol{A'}$: predicted answer,
$\boldsymbol{Q'}$: generated question.
Overlapping words in the questions are \ul{underlined}.}
\label{tb:example}
\end{table*}

\section{Introduction}
Question answering (QA) for machine reading comprehension is a central task in natural language understanding, which requires a model to answer questions given textual contexts.
Pretrained language models have been successfully applied to QA and achieve scores higher than those of humans on benchmark datasets such as SQuAD \cite{rajpurkar-etal-2016-squad}.
However, QA models have been demonstrated to exploit unintended dataset biases instead of the intended solutions, and lack robustness to challenge test sets whose distributions are different from those of training sets \cite{jia-liang-2017-adversarial,sugawara-etal-2018-makes,gan-ng-2019-improving,ribeiro-etal-2019-red}, which could be a serious problem in real-world applications.

Question generation (QG) has also been extensively studied to augment QA datasets \cite{du-etal-2017-learning,du-cardie-2018-harvesting}.
It is demonstrated that QG can improve not only the in-domain generalization but also the out-of-distribution generalization capability of QA models \cite{zhang-bansal-2019-addressing,lee-etal-2020-generating,shinoda-etal-2021-improving}.
In other areas, data augmentation techniques have been successfully used to reduce dataset biases and increase the performance of machine learning models on under-represented samples in vision \cite{McLaughlin,7797091} and language \cite{zhao-etal-2018-gender,zhou-bansal-2020-towards}.
Thus, we assume that QG is useful to debias QA models and improve its robustness by augmenting QA datasets.
However, it has not been fully studied whether existing QG models can contribute to debiasing QA models (i.e., improve the robustness of QA models to under-represented questions).

In this study, we focus on question--context lexical overlap, inspired by the findings presented in \citet{sugawara-etal-2018-makes}.
Their work revealed that questions having low lexical overlap with context tend to require reasoning skills rather than superficial word matching, and existing QA models are not robust to these questions (Table \ref{tb:example}).
To see if data augmentation with recent neural QG models can improve the robustness to those questions, we analyze the performance of BERT \cite{bert} trained on SQuAD v1.1 \cite{rajpurkar-etal-2016-squad} augmented with them.
Our analysis reveals that data augmentation with neural QG models frequently sacrifices the QA performance of the BERT-base model on questions with low lexical overlap, while improving that on questions with high lexical overlap.
We conjecture that this is because neural QG models frequently generate questions with high lexical overlap as indicated in Table \ref{tb:example}.
This behavior can be interpreted as a consequence of the recent QG models pursuing higher average BLEU scores on SQuAD, which inherently contains reference questions with high lexical overlap, by copying many words from contexts to generate questions.
By doing so, QG models can amplify the lexical overlap bias in the original dataset.

To address the performance degradation, we use a simple data augmentation approach using synonym replacement to generate questions with low question--context lexical overlap.
We found that the proposed approach not only debiases the dataset but also improves the QA performance on questions with low lexical overlap with only 70k synthetic examples, whereas conventional neural QG approaches use more than one million synthetic examples.

In summary, our contributions are as follows:
\begin{itemize}
    \item We found that not only QA but also QG models are biased in terms of question--context lexical overlap; that is, QG models fail to generate questions with low lexical overlap (\S\ref{sec:reevaluate}).
    \item We discovered that data augmentation using recent neural QG models does not contribute to debias QA datasets; rather, it frequently degrades the QA performance on questions with low lexical overlap, while improving that on questions with high lexical overlap (\S\ref{sec:da}).
    \item We demonstrated that the proposed simple data augmentation approach using synonym replacement (\S\ref{sec:method}) for augmenting questions with low lexical overlap is effective to improve QA performance on questions with low lexical overlap with only 70k synthetic examples (\S\ref{sec:da}), while preserving or slightly hurting the overall accuracy.
\end{itemize}

\section{Revisiting the QA and QG Performance in Terms of Question--Context Lexical Overlap}
\label{sec:reevaluate}
In this paper, we denote question--context lexical overlap as QCLO.
We define QCLO as the ratio of the overlapping words between question $Q$ and context $C$ to the total number of words in question.\footnote{When computing lexical overlap, we do not exclude stop words because even overlapping stop words are important cues to determine the correct answer.}
Precisely, QCLO is calculated as
\begin{align}
\label{eq:lo}
\textrm{QCLO} = \frac{|\boldsymbol{Q} \cap \boldsymbol{C}|}{|\boldsymbol{Q}|}.
\end{align}
The second example in Table \ref{tb:example} indicates a question with lower QCLO is neither answered nor generated correctly by neural models.
To investigate this phenomenon, we first analyze the QA and QG performance in terms of QCLO.

\paragraph{Experimental setups}
For QA, we use the fine-tuned BERT-base and -large models \cite{bert}.
For QG, we use SemanticQG \cite{zhang-bansal-2019-addressing}.\footnote{We used the ELMo+QPP\&QAP \cite{zhang-bansal-2019-addressing} model for QG.}
For the dataset, we use the SQuAD-Du dataset; the train, dev, and test split of SQuAD v1.1 \cite{rajpurkar-etal-2016-squad} proposed by \citet{du-etal-2017-learning}, which we denote as \squaddu{train}, \squaddu{dev}, \squaddu{test}, respectively.
This split is commonly used in the QG literature \cite{du-cardie-2018-harvesting,zhang-bansal-2019-addressing} because the original test set is not released.
The numbers of question, answer and context triples in \squaddu{train}, \squaddu{dev}, and \squaddu{test}, are 76k, 11k, and 12k, respectively.

\begin{figure}[t]
    \centering
    \begin{tabular}{rl}
         \includegraphics[clip,width=3.5cm]{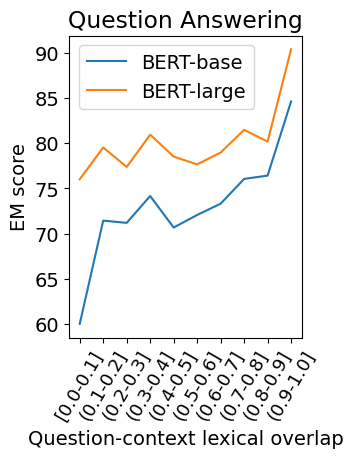} & \includegraphics[clip,width=3.5cm]{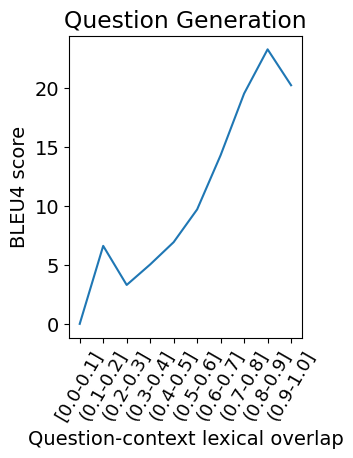} \\
    \end{tabular}
    \caption{Exact match (EM) score of BERT models and BLEU-4 score of SemanticQG \cite{zhang-bansal-2019-addressing} on the test set of SQuAD-Du for each range of Question--Context Lexical Overlap (QCLO). See Eq. \ref{eq:lo} for the definition of QCLO. Both the QA and QG models degrade the scores on questions with low QCLO.}
    \label{fig:qa-qg}
\end{figure}

\paragraph{Results}
We show the result in Figure \ref{fig:qa-qg}.
This indicates that the performance of the BERT models on the questions with lower QCLO is degraded compared to the questions with higher QCLO.
For QG, the BLEU-4 score \cite{bleu} is highly correlated with QCLO, which means that the model fails to generate questions with low QCLO accurately.

\begin{figure}[t]
\centering
\includegraphics[clip,width=8cm]{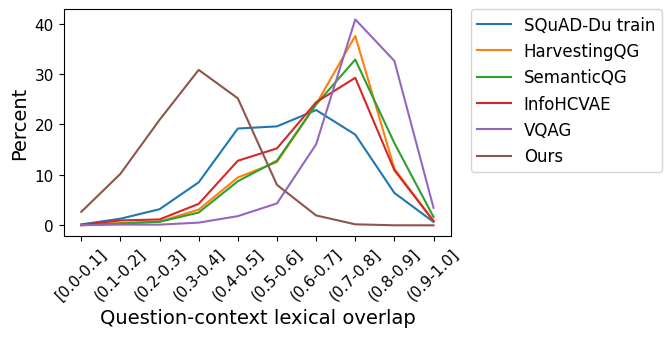}
\caption{The percentages of questions in the datasets, SQuAD-Du \cite{du-etal-2017-learning}, HarvestingQG \cite{du-cardie-2018-harvesting}, SemanticQG \cite{zhang-bansal-2019-addressing}, InfoHCVAE \cite{lee-etal-2020-generating}, VQAG \cite{shinoda-etal-2021-improving}, and ours (\S\ref{sec:method}), for each range of QCLO. While neural question generation models are biased towards generating questions with high QCLO, ours can generate questions with low QCLO.
\label{fig:lo}
}
\end{figure}

We also show the distributions in terms of QCLO of questions generated by recent neural QG models (HarvestingQG \cite{du-cardie-2018-harvesting}, SemanticQG \cite{zhang-bansal-2019-addressing}, InfoHCVAE \cite{lee-etal-2020-generating}, and VQAG \cite{shinoda-etal-2021-improving}) in Figure \ref{fig:lo}.
This indicates that all the QG models are biased towards generating questions with higher QCLO than \squaddu{train}, which is used to train those QG models.

Based on the result, we suspect that when neural QG is used to augment a QA dataset, the degraded QG performance on questions with low QCLO could exacerbate the degraded QA performance.
Our experiments in \S\ref{sec:da} show that this is often true.
We hypothesize that this is caused by the strong tendency of neural QG models to generate questions with high QCLO as shown in Figure \ref{fig:lo}.

\section{Method}
\label{sec:method}
We assume that if we augment questions with low QCLO unlike existing neural QG approaches, the robustness of QA models to questions with low QCLO can be improved.
In this section, we describe the proposed method for generating questions with low QCLO.
We extend the idea of synonym replacement used in \cite{wei-zou-2019-eda} to reduce the lexical overlap.
The proposed method is as follows:
\begin{enumerate} % [leftmargin=1.5em]
    \item List all the overlapping words between question and context.
    \item Replace every word in the listed words other than predefined stop words with one of its synonyms chosen randomly from WordNet \cite{wordnet}, and obtain a synthetic question.
    \item If the lexical overlap decreases after synonym replacement, add the synthetic question to our dataset; if not, discard the question.
\end{enumerate}
After repeating this procedure once for every ground-truth question in the training set, we obtain 70k synthetic questions with significantly lower lexical overlap, as indicated in Figure \ref{fig:lo} (ours).
For example,
\textit{What is heresy mainly at odds with?} is converted into \textit{What is heterodoxy mainly at odds with?}, and
\textit{How many documents remain classified?} is converted into
\textit{How many text file remain classified?}.
Because \textit{heterodoxy}, \textit{text}, and \textit{file} do not appear in the contexts, the lexical overlap is reduced in each example.

It is worth mentioning a couple of limitations of our method.
First, synonym replacement may slightly change the meaning of questions depending on the context. Second, our approach relies on the assumption that annotated questions are available, which makes it impossible to apply to unlabeled passages.

\section{Experiments}
\label{sec:da}
To determine the effect of data augmentation on improving the QA model robustness to questions with low QCLO, we conducted experiments with several QG approaches.

\paragraph{Dataset}
We used the SQuAD-Du dataset as in \S\ref{sec:reevaluate}.
Considering the QCLO statistics of SQuAD displayed in Figure \ref{fig:lo}, we split \squaddu{dev} and \squaddu{test} into \textbf{Easy} and \textbf{Hard} subsets that contain questions with QCLO greater than 0.3, and the others, respectively.
Our Easy and Hard subsets offered concise, yet sufficient, evaluation in terms of QCLO.

\paragraph{Baselines}
We adopted the following four baselines that use neural QG models for data augmentation.
\begin{itemize} % [leftmargin=1em]
    \item \textbf{HarvestingQG} \cite{du-cardie-2018-harvesting} generates question--answer pairs from 10,000 top-ranking Wikipedia articles with neural answer extraction and question generation.\footnote{\url{https://github.com/xinyadu/harvestingQA}} The size is 1.2 million.
    \item \textbf{SemanticQG} \cite{zhang-bansal-2019-addressing} is a QG model that uses reinforcement learning to generate semantically valid questions.
    Following this work, we generated questions using the publicly available model\footnote{\url{https://github.com/ZhangShiyue/QGforQA}} from the same context--answer pairs as HarvestingQG. The size is 1.2 million.
    \item \textbf{InfoHCVAE} \cite{lee-etal-2020-generating} is a question--answer pair generation model based on conditional variational autoencoder with mutual information maximization. We trained this model on \squaddu{train}, and then generated 50 questions and answers from each context in \squaddu{train}. The size is 824k.
    \item \textbf{VQAG} \cite{shinoda-etal-2021-improving} is a question--answer pair generation model based on conditional variational autoencoder with explicit KL control. We used the publicly available dataset.\footnote{\url{https://github.com/KazutoshiShinoda/VQAG}} The size is 432k.
\end{itemize}
The distributions of the lexical overlap of these datasets are presented in Figure \ref{fig:lo}.
We indicate that these methods are more biased towards high lexical overlap than \squaddu{train}, which was used as the training set for these QG models.

\paragraph{Experimental Setups}
As in our previous experiment (\S\ref{sec:reevaluate}), we used BERT-base and -large models, whose total number of parameters are 110M and 340M, respectively.
\citet{dhingra-etal-2018-simple} proposed to pretrain a QA model using synthetic data composed of cloze-style questions and then fine-tune it on the ground-truth data.
We adopted the pretrain-and-fine-tune approach for the neural QG approaches, which generated over 1.2 million questions.
However, as discussed by \citet{zhang-bansal-2019-addressing}, we observed that when the size of the synthetic data was small or similar to the ground-truth data, a performance gain could not be obtained by the pretrain-and-fine-tune approach.
Thus, for the proposed approach, which generated 70k questions, we fine-tuned QA models on the ground-truth data randomly mixed with the generated data.
We used the Hugging Face's implementation of BERT \cite{wolf2019huggingface}.
We use the Adam \cite{kingma2014adam} optimizer with epsilon set to 1e-8.
The batch size was 32 for all the settings.
In both the pretraining and fine-tuning procedure, the learning rate decreased linearly from 3e-5 to zero.
We train the QA models for one epoch for pretraining with synthetic data and two epochs for fine-tuning with \squaddu{train}.

\begin{table*}[tbp]
    \centering
    \scalebox{0.85}{
    \begin{tabular}{cc|ccc|ccc}
    \toprule
     & & \multicolumn{3}{c|}{\squaddu{dev} (EM/F1)} & \multicolumn{3}{c}{\squaddu{test} (EM/F1)} \\
    \midrule
    Model & Train Source & Hard & Easy & ALL & Hard & Easy & ALL \\
    \midrule
    \multirow{6}{*}{base} & \squaddu{train} & 72.31/81.11 &80.74/88.39 & 80.35/88.05& 70.88/81.99 &73.22/84.75 & 73.06/84.57\\
    & + HarvestingQG & 70.25/78.27 &80.06/87.62 & 79.60/87.19& 69.28/79.92 &73.15/84.20 & 72.90/83.93\\
    & + SemanticQG & 70.45/80.25 &81.70/89.08 & 81.17/88.67& 71.68/82.49 &\textbf{74.39}/\textbf{85.59} & \textbf{74.21}/\textbf{85.39}\\
    & + InfoHCVAE & 72.05/80.66 &81.79/\textbf{89.35} & 81.34/\textbf{88.95}& 73.47/\textbf{83.91} &73.50/85.08 & 73.48/84.99\\
    & + VQAG        	& 73.29/82.04 & \textbf{81.88}/88.93 & \textbf{81.48}/88.62& 71.60/83.07 &73.79/85.23 & 73.63/85.08\\
    \cmidrule{2-8}
    & + Ours & \textbf{73.50}/\textbf{82.81} &80.34/87.81 & 80.02/87.58& \textbf{73.60}/83.49 &73.08/84.41 & 73.11/84.34\\
    \midrule
    \multirow{6}{*}{large} & \squaddu{train} & 78.72/87.71 &\textbf{87.06}/\textbf{93.23} & \textbf{86.67}/\textbf{92.98}& 77.93/87.84 &\textbf{79.33}/\textbf{89.88} & \textbf{79.24}/\textbf{89.74}\\
    & + HarvestingQG & 79.13/86.92 &85.55/92.12 & 85.26/91.88& 76.99/86.61 &77.58/88.28 & 77.54/88.17\\
    & + SemanticQG & 79.96/87.73 &85.90/92.57 & 85.62/92.35& 76.99/87.29 &77.82/88.68 & 77.77/88.59\\
    & + InfoHCVAE & 77.85/86.44 &85.25/92.15 & 84.91/91.89& 76.00/87.55 &78.02/88.90 & 77.87/88.80\\
    & + VQAG      	& 79.50/87.55 &86.68/93.01 & 86.35/92.76& 77.33/87.70 &78.98/89.36 & 78.86/89.25\\
    \cmidrule{2-8}
    & + Ours & \textbf{81.37}/\textbf{88.33} & 86.49/92.78 & 86.25/92.57& \textbf{78.40}/\textbf{88.52} &77.94/89.00 & 77.96/88.97\\
    \bottomrule
    \end{tabular}
    }
    \caption{QA performance with data augmentation. EM/F1 scores on the Hard (where ${\rm QCLO} \leq 0.3$) and Easy (where ${\rm QCLO} > 0.3$) subsets, and the whole set of \squaddu{dev} and \squaddu{test} are reported.
    }
    \label{tb:main}
\end{table*}

\paragraph{Results}
The results of the data augmentation are displayed in Table \ref{tb:main}. In all the settings, the proposed approach achieved the best EM score on the Hard subset.
Notably, the proposed method significantly improved the performance by \textbf{2.72 (EM)} / \textbf{1.50 (F1)} points using BERT-base on the Hard subset in the test set, while maintaining the overall scores compared to the no data augmentation baseline.
This improvement indicates that the proposed approach for debiasing the dataset in terms of QCLO is helpful for addressing the performance degradation.
However, the proposed approach degraded the scores on the Easy subsets when using BERT-large.
Addressing the trade-off between the scores in the Hard and Easy subsets using BERT-large is future work.

When using BERT-base, the neural QG baselines except for HarvestingQG improved the scores on the Easy subset; however, the baselines except for InfoHCVAE often degraded the scores on the Hard subset.
This could be due to the tendency to generate questions with high QCLO (Figure \ref{fig:lo}).

When using BERT-large, the QG approaches often fail to improve the scores in both the Hard and Easy subsets.
Generating useful examples for a larger model is more challenging than for a smaller one according to these results.
Utilizing pretrained language models for QG may be useful given the fact that only RNNs are used in all the baseline QG methods in our experiments.

HarvesingQG was not effective in almost all the settings.
Comparing its scores with those of SemanticQG, which used the same context--answer pairs as HarvestingQG, some feature of generated questions other than lexical overlap appeared to be critical in improving the QA scores on the Easy subset, because the distributions of QCLO of two synthetic datasets were similar to each other (see Figure \ref{fig:lo}).

For further boosting the overall average score, we can make an ensemble prediction using the best performing models in the Easy and Hard subsets, although improving the overall scores is not the main focus in this paper.
The performance gains were positive but not very significant in our case.
We leave utilizing the ensemble prediction to address the performance trade-off to future work.

\begin{table*}[tbp]
\centering
\begin{tabular}{p{15.5cm}}
\toprule
Besides earning a reputation as a respected \textcolor{red}{entertainment} \textit{\scriptsize (Original, InfoHCVAE)} device, the iPod has also been accepted as a \textbf{business} \textit{\scriptsize (Ours)} device.
Government departments, major institutions and international organisations have turned to the iPod line as a delivery mechanism for business communication and training, such as \textcolor{red}{the Royal and Western Infirmaries} \textit{\scriptsize (HarvestingQG, SemanticQG, VQAG)} in Glasgow, Scotland, where iPods are used to train new staff.\\
--- Aside from recreational use, \ul{in} what other arena \ul{have} \ul{iPods} found use? \hfill (QCLO: \colorbox{red!50}{0.29})\\
\midrule
In \textbf{2010} \textit{\scriptsize (Ours)}, a number of workers committed suicide at a Foxconn operations in China. Apple, HP, and others stated that they were investigating the situation. Foxconn guards have been videotaped beating employees. Another employee killed himself in \textcolor{red}{2009} \textit{\scriptsize (Original, HarvestingQG, SemanticQG, VQAG)} when an Apple prototype went missing, and claimed in messages to friends, that he had been beaten and interrogated.\\
--- \ul{In} what year did Chinese \ul{Foxconn} emplyees* kill themselves? (*: annotator's typo) \hfill (QCLO: \colorbox{red!50}{0.2})\\
\midrule
The BBC began its own regular television programming from the basement of Broadcasting House, London, on \textcolor{red}{22 August 1932} \textit{\scriptsize (HarvestingQG, SemanticQG)}. The studio moved to larger quarters in 16 Portland Place, London, in \textbf{February 1934} \textit{\scriptsize (Original, Ours)}, and continued broadcasting the 30-line images, carried by telephone line to the medium wave transmitter at Brookmans Park, until 11 September 1935, by which time advances in all-electronic television systems made the electromechanical broadcasts obsolete.
\\
--- When did \ul{the} \ul{BBC} first change studios? \hfill (QCLO: \colorbox{red!50}{0.25})\\
\midrule
\textcolor{red}{Peyton Manning} \textit{\scriptsize (VQAG)} became the first quarterback ever to lead two different teams to multiple Super Bowls. He is also the oldest quarterback ever to play in a Super Bowl at age \textcolor{red}{39} \textit{\scriptsize (Original, Ours)}. The past record was held by \textbf{John Elway} \textit{\scriptsize (HarvestingQG, SemanticQG, InfoHCVAE)}, who led the Broncos to victory in Super Bowl XXXIII at age 38 and is currently Denver's Executive Vice President of Football Operations and General Manager.\\\
--- Prior \ul{to} \ul{Manning}, \ul{who} \ul{was} \ul{the} \ul{oldest} \ul{quarterback} \ul{to} \ul{play} \ul{in} \ul{a} \ul{Super} \ul{Bowl}?
\hfill (QCLO: \colorbox{green!60}{0.88})\\
\midrule
Despite being relatively unaffected by \textcolor{red}{the embargo} \textit{\scriptsize (Original, HarvestingQG, VQAG, Ours)}, the UK nonetheless faced an oil crisis of its own - \textbf{a series of strikes by coal miners and railroad workers} \textit{\scriptsize (SemanticQG, InfoHCVAE)} over the winter of 1973–74 became a major factor in the change of government. Heath asked the British to heat only one room in their houses over the winter. The UK, Germany, Italy, Switzerland and Norway banned flying, driving and boating on Sundays. Sweden rationed gasoline and heating oil. The Netherlands imposed prison sentences for those who used more than their ration of electricity.\\
--- What caused \ul{UK} \ul{to} have \ul{an} \ul{oil} \ul{crisis} \ul{in} \ul{its} \ul{own} country? \hfill (QCLO: \colorbox{green!60}{0.62})\\
\bottomrule
\end{tabular}
\caption{Illustrative predictions on \squaddu{dev} and \squaddu{test} by a BERT-base model trained on \squaddu{train} (Original), +HarvestingQG, +SemanticQG, +InfoHCVAE, +VQAG, and +Ours. The ground truth answers are in \textbf{bold}. The incorrectly predicted answers are written in \textcolor{red}{red}.
The QA models that predict them are written in \textit{italics}.
The overlapping words in the questions are \ul{underlined}. Question--context lexical overlap (QCLO) is given in parentheses.}
\label{tb:case}
\end{table*}

\section{Qualitative Analysis}
To demonstrate the effect of the baseline QG models and proposed method qualitatively, we present examples in both the Hard (QCLO $\leq$ 0.3) and Easy (QCLO $>$ 0.3) subsets in Table \ref{tb:case}.
The first two examples show that only the QA model trained with the proposed method could correctly answer the questions.
Answering the questions in these examples required a knowledge of synonyms, such as ``recreational'' vs. ``entertainment,'' ``besides'' vs. ``aside from,'' ``employees'' vs. ``workers,'' and ``kill oneself'' vs. ``commit suicide.''
These examples imply that the proposed data augmentation method based on synonym replacement enabled the QA model to acquire knowledge regarding synonyms.
This kind of reasoning beyond superficial word matching is indispensable for QA systems to achieve human-level language understanding.

The third example in Table \ref{tb:case} displays an example where data augmentation using the neural QG models made the original prediction incorrect.
This example implies that current QG models may harm the robustness of QA models to questions with low QCLO.
As \citet{geirhos_shortcut_2020} discussed, if QG models just amplify the dataset bias, QA models could learn dataset-specific solutions (i.e., shortcuts) and fail to generalize to challenge test sets.

In contrast, the fourth and fifth examples in Table \ref{tb:case} display examples in the Easy subset where data augmentation with neural QG models is beneficial, while the original and proposed models fail to answer them correctly.
These examples require multiple-sentence reasoning, i.e., one has to read and understand multiple sentences to answer these questions. 
This observation implies that some under-represented features (e.g., multiple-sentence reasoning \cite{rajpurkar-etal-2016-squad}) exist even in the Easy subset, and the existing neural QG models might amplify such features (possibly by copying many words from multiple sentences to formulate questions) and make it easy to capture them.
Investigating what kind of features are learned by using data augmentation with neural QG models in more detail is future work.

\section{Related Work}
\paragraph{The Robustness of QA models}
Pretrained language models such as BERT \cite{bert} have surpassed the human score on the SQuAD leaderboard.\footnote{\url{https://rajpurkar.github.io/SQuAD-explorer/}}
However, such powerful QA models have been shown to exhibit the lack of robustness.
A QA model that is trained on SQuAD is not robust to paraphrased questions \cite{gan-ng-2019-improving}, implications derived from SQuAD \cite{ribeiro-etal-2019-red}, questions with low lexical overlap \cite{sugawara-etal-2018-makes}, and other QA datasets \cite{yogatama2019learning,talmor-berant-2019-multiqa,sen-saffari-2020-models}.
\citet{ko-etal-2020-look} showed that extractive QA model can suffer from positional bias and fail to generalize to different answer positions.

The lack of robustness demonstrated in these studies can be explained by shortcut learning of deep neural networks \cite{geirhos_shortcut_2020}.
A high score on an in-distribution test set can be achieved by just exploiting unintended dataset biases \cite{levesque-2014-best-behaviour}.
Therefore, evaluating QA models only on an in-distribution test set is not enough to evaluate the robustness of the QA models.

\paragraph{Question Generation for Question Answering}
QG has been studied extensively in order to augment QA datasets and boost the QA performance, which has been evaluated primarily on SQuAD \cite{du-etal-2017-learning,zhou-etal-2017-neural,yang-etal-2017-semi,zhang-bansal-2019-addressing}.
Question answer pair generation, which consists of answer candidate extraction and QG, has been also received attention because question-worthy answers for the input of QG are not freely available \cite{du-cardie-2018-harvesting,lee-etal-2020-generating,shinoda-etal-2021-improving}.
The de facto standard of QG models is to utilize a copy mechanism \cite{Gu16,Gulcehre16}.
The tendency of QG models to copy words from textual contexts as indicated in Figure \ref{fig:lo} is partially due to this copy mechanism.
While the existing QG works have increased the BLEU scores on SQuAD\footnote{\url{http://aqleaderboard.tomhosking.co.uk/squad}} and successfully generated fluent questions in terms of human scores, the bias regarding lexical overlap in QG has not received sufficient attention.

\paragraph{Data Augmentation and Dataset Bias}
Data augmentation has been widely used in other domains to reduce dataset biases such as the background bias in person re-identification \cite{McLaughlin}, the gender bias in coreference resolution \cite{zhao-etal-2018-gender}, and the lexical bias in natural language inference \cite{zhou-bansal-2020-towards}.
These works repeated training examples or added synthetic data to increase under-represented samples and reduce the imbalance in a training set.
Our proposed approach has the same motivation as these works.

On the other hand, data augmentation can unintentionally introduce or amplify dataset bias.
Back-translation \cite{sennrich-etal-2016-improving}, which is the common data augmentation approach for machine translation, can introduce the translationese bias.
That is, machine translation systems trained with back-translation, compared to ones without back-translation, can enhance the BLEU scores when the input is translationese (i.e., human-translated texts) but harm the BLEU scores when the input is naturally occurring texts \cite{edunov-etal-2020-evaluation,marie-etal-2020-tagged}.
This phenomenon is analogous to the observation in our work, where we demonstrated that SQuAD QG models are biased towards generating questions with high QCLO, and this tendency can harm the QA performance on questions with low QCLO while improving that on questions with high QCLO.

\section{Conclusion}
We demonstrated that not only QA models but also QG models are biased in terms of the question--context lexical overlap.
To determine the influence of the bias, we analyzed the QA performance with data augmentation using the recent QG models.
We demonstrated that they frequently degraded the QA performance on questions with low lexical overlap, while improving that on questions with high lexical overlap when using BERT-base.
To address this problem, we designed a simple approach using synonym replacement to debias a QA dataset.
We demonstrated that the proposed approach improved the QA performance on questions with low lexical overlap while maintaining or slightly degrading the overall scores with only 70k synthetic examples.

Our results suggest that future research in QG for data augmentation should exercise caution to prevent the amplification of dataset bias in terms of lexical overlap.
In addition, what features are learned by data augmentation with neural QG models is worth to be explored in more detail to clarify what is improved and what is not improved by QG.
It is also worth investigating whether our findings still hold in other QA datasets where annotated questions have lower lexical overlap than those in SQuAD.

\section*{Acknowledgements}
We would like to thank the anonymous reviewers for their detailed and valuable comments.
This work was supported by NEDO SIP-2 ``Big-data and AI-enabled Cyberspace Technologies,'' and JSPS KAKENHI Grant Numbers 21H03502, 20K23335.

\bibliography{anthology,custom}
\bibliographystyle{acl_natbib}

\appendix

\end{document}